\documentclass[conference]{IEEEtran}
\usepackage{cite}
\usepackage{graphicx}
\usepackage{amsmath}
\usepackage{algorithm2e}
\usepackage{url}
\begin{document}
\bstctlcite{IEEEexample:BSTcontrol}
\RestyleAlgo{ruled} 
\title{Federated Learning for Water Consumption Forecasting in Smart Cities}

\author{
\IEEEauthorblockN{Mohammed El Hanjri\IEEEauthorrefmark{3}, Hibatallah Kabbaj\IEEEauthorrefmark{3}, Abdellatif Kobbane\IEEEauthorrefmark{3}, Amine Abouaomar\IEEEauthorrefmark{1}}
\IEEEauthorblockA{\IEEEauthorrefmark{3}ENSIAS, Mohammed V University in Rabat, Morocco.}
\IEEEauthorblockA{\IEEEauthorrefmark{1}School of Science and Engineering, Al Akhawayn University in Ifrane, Morocco.}
}

\maketitle

\begin{abstract}
Water consumption remains a major concern among the world's future challenges. For applications like load monitoring and demand response, deep learning models are trained using enormous volumes of consumption data in smart cities. On the one hand, the information used is private. For instance, the precise information gathered by a smart meter that is a part of the system's IoT architecture at a consumer's residence may give details about the appliances and, consequently, the consumer's behavior at home. On the other hand, enormous data volumes with sufficient variation are needed for the deep learning models to be trained properly. This paper introduces a novel model for water consumption prediction in smart cities while preserving privacy regarding monthly consumption. The proposed approach leverages  federated learning (FL) as a machine learning paradigm designed to train a machine learning model in a distributed manner while avoiding sharing the users data with a central training facility. In addition, this approach is promising to reduce the overhead utilization through decreasing the frequency of data transmission between the users and the central entity. Extensive simulation illustrate that the proposed approach shows an enhancement in predicting water consumption for different households.

\end{abstract}

\IEEEpeerreviewmaketitle

\section{Introduction}
In order to develop smart cities, load forecasting is a crucial step.  Long-term consumption forecasts of various vital resources are essential for infrastructure management and planning \cite{9220170, taik2020icc}. In addition, short and medium term consumption forecasts are critical for system performance. The daily operational efficiency of water distribution, in particular, requires accurate short-term forecasting, which relies on the collection and analysis of large volumes of data from the various domains\cite{taik2021smartgrid}. However, the forecasting of individual short-term loads is a challenging task to accomplish due to the high volatility of user profiles. Indeed, the water demand of a household or building is strongly influenced by the behavior of its residents, which is too stochastic and therefore difficult to anticipate.

To predict load and consumption, different approaches have been proposed \cite{111111}, including linear and dynamic programming, heuristic methods, game theory, and fuzzy methods \cite{111112}. The problem is that existing centralized and decentralized approaches do not consider online solutions for large-scale real-world databases, which is usually the case for residential consumption \cite{111113, abouaomar2022federated}. Due to large amounts of data from water meters, these approaches generate notable slowdowns, which are not optimal for real-time applications. In the era of big data, many machine learning methods are emerging as suitable for overcoming this limitation by automatically exploiting, controlling, and optimizing load prediction models \cite{111114}. This can be achieved by applying successive transformations to historical data to train robust machine learning models to accommodate the high uncertainty of water patterns, and control demand based on load patterns and market cost signals.

By utilizing the caching capabilities coupled with the Fog characteristics, the authors provide a green method to enhance performance of the energy efficiency in a fog computing environment and Internet of Everything (IoE) devices. To handle distributed networking, processing, and storage resources, they make use of fog computing. As content requesters with limited resources, IoT devices will make use of cutting-edge caching techniques to deliver high-quality services that demand a lot of computational power and a high throughput \cite{8422154, assila2018achieving, gmira2015new}.

Recently, the machine learning community proposed Federated Learning (FL) as a novel on-device method to address privacy issues while also growing the number and variety of data sets \cite{mcmahan2017communication}.
A decentralized machine learning method called federated learning allows any device to train a central model without delivering any data \cite{9220170, abouaomar2022federated}. The model is initially initialized by the server either randomly or using data that is readily accessible to the public \cite{chen2021distributed}.
The model is then given to a group of clients that have been chosen at random, where it will be trained locally using their data.
Each client updates the model's weights, which are then communicated to the server, averaged, and used to update the overall model.
This process will be continued until the global model stabilizes \cite{kairouz2021advances}.

This paper's primary goal is to assess how well the Federated Learning approach worked for the STLF challenge on home water use.
For time series forecasting, we use a deep neural network called Long Short Term Memory (LSTM) \cite{hochreiter1997long}, which makes predictions about future measurements of the water consumption of the home based on past ones. We investigate a set of structures with comparable characteristics (geographical location, type of building). Federated learning is carried out using home grid Edge technology \cite{9240934, 9326402}. Edge equipment, such as smart meters or more sophisticated devices, is frequently installed at the end of the water distribution system to serve as a smart interface between the customer and the water supply. We made the following contributions to this effort, in brief: Using Edge technology in smart cities, We first suggest an enabling architecture for FL, then we simulate the potential accuracy gain of FL and last we use numerical findings to simulate the potential network load gain. We also include in these contributions the improvement in privacy made available by decentralization and Edge computing.

The rest of this paper is organized as follows, with section II discussing related works with a focus on load forecasting and privacy. The system model is described in Section III, while Section IV presents the simulation and numerical findings. Section V serves as the paper's conclusion. \\

\section{Related Work}

Deep neural networks, especially in particular Long-short term memory (LSTM), have been employed extensively in recent research to address the difficulty of short-term load forecasting.
However, the results fall short of the required level of precision in terms of Mean Average Percentage Error (MAPE) and Root Mean Square Error (RMSE), despite benchmarks demonstrating LSTM's capability in comparison to other methods \cite{bouktif2018optimal}.  The authors of \cite{marino2016building} suggest using a sequence-to-sequence LSTM version of LSTM to increase forecasting accuracy. This variant performs better for data with a one-minute resolution but does not significantly outperform regular LSTM for data with an hour resolution. Additionally, some writers \cite{almalaq2018evolutionary} employ the evolutionary algorithm to solve the challenge of selecting the optimal LSTM network because they view it as a hyperparameter tuning problem. 

According to some other studies, the issue is actually with data-driven forecasting models capacity to generalize, not just with neural network architecture.
In fact, when applied to fresh datasets, several of the presented models accuracy decreases \cite{bouktif2018optimal}. 

Grouping data from several clients is another method for enhancing the training data. By grouping people with comparable profiles, the authors of \cite{stephen2015incorporating} are able to lower the variation of uncertainty within groups. To solve the overfitting issue, authors in \cite{shi2017deep} suggest a pooling strategy that broadens the diversity of the data. However, these approaches are very centralized and prone to privacy problems. 

In the proposed study, we recommend using the federated learning strategy to carry out client selection and neural network training at the edge. This will allow the use of data to build a reference model without sacrificing resident privacy.
In order to do this, the Home Area Network (HAN) Edge Equipment is utilized.

\section{System Model}


We suggest the network architecture shown in Figure 1, which has a Multi-access Edge Computing (MEC) server and users as its two primary parts. The users are buildings having Edge equipment made up primarily of smart meters and other HAN devices. For STLF, FL is utilized to create a global model based on LSTM.
The MEC server orchestrates the training rounds, which are carried out by the users using information about their local water consumption. We go into greater detail regarding LSTM and its use in forecasting, as well as FL and its use in our system model, in this part. \\

\begin{figure*}[h!]
    \centering
    \includegraphics[width=0.8\linewidth]{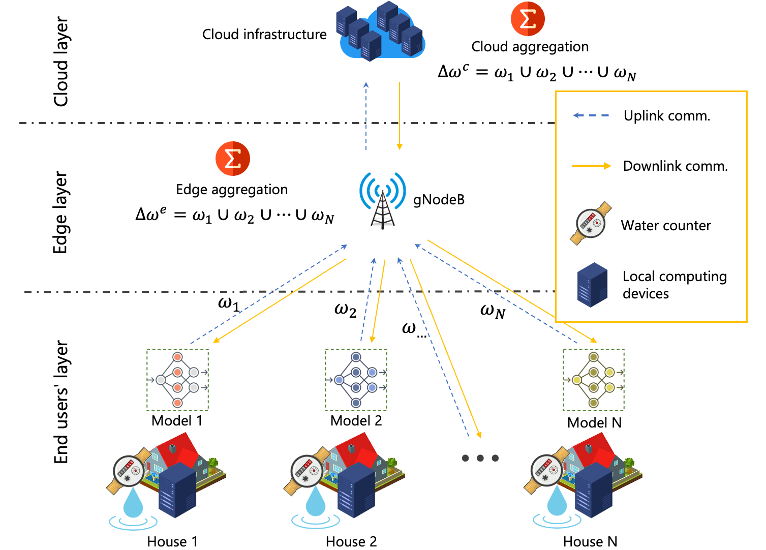}
    \caption{Network architecture and components}
    \label{fig:sm}
\end{figure*}

\subsection{Time series forecasting using LSTM}
In this work, the time series forecasting technique with LSTM is used to predict the future water load. A time series is an organized collection of equally spaced data sets that shows how a certain variable has changed over time. By modeling the connections between the points of the current data and the points of the past, time series forecasting is made possible, but the precision of the forecasts is greatly influenced by the model that is used and the caliber of the historical data points. \\
Recurrent neural networks (RNNs) like LSTM are more effective than typical RNNs and fundamentally different from conventional feedforward RNNs. The strength of LSTM is sequence learning. It can resolve vanishing and exploding gradient difficulties, which are frequent in RNNs, and establish temporal correlations between earlier data points and the current situation. Gradient exploding relates to the opposite process, whereas gradient vanishing describes how the norm of the gradient for long-term components gets less, leading weights to never change at lower levels. This is made possible by two of its essential elements: the memory cell, which is utilized to recall significant prior events, and the gates, which control the information flow. The input gate, output gate, and forget gate are the three gates that make up an LSTM. During the learning process, they acquire the skill of clearing the memory of irrelevant data.
Utilizing LSTM and its derivatives, nearly all state-of-the-art outcomes in sequence learning are accomplished.
For residential STLF, it is anticipated that the LSTM network will be able to track the states of the inhabitants, abstract some states from the provided consumption profile, and forecast future consumption using the learned data.

\subsection{Federated Learning}

A type of machine learning called federated learning involves distributing the majority of the training process across various clients, or computer devices. On mobile device keyboards for next word prediction, Google first put out and executed the idea \cite{hard2018federated}. For many situations, this strategy is ideal: 1) In highly distributed systems, when there are orders of magnitude more devices than data center nodes, 2) When data privacy is important, and 3) When data is huge in comparison to model updates. In cases when datasets are uneven or not dispersed evenly, federated learning has been shown to be quite effective.\\
Following is an example of federated learning iteration:
First, the current model is given to each client in a selected subset. In this instance, clients are housed using Edge equipment (e.g. smart meters). On locally stored data, a few clients choose to compute Stochastic Gradient Descent (SGD) updates. A server then integrates the client updates to produce a new global model. A different group of clients is sent the updated model. Until the necessary forecast accuracy is attained, this process is repeated. Algorithm 1 describes the procedures in detail.\\
The server employs the Federated Averaging algorithm to integrate the client updates \cite{mcmahan2017communication}.
The starting global model is first randomly initialized or trained beforehand using publicly accessible data. A subset $S$ of clients with sufficient data records and diverse enough consumption loads to augment the training data are sent a global model $\omega_t$ by the server in each training round $t$. This requirement was introduced to make sure we have enough variety among the data points to accurately depict the occupants' typical consumption. Each client $s$ in the subset then employs $m_s$ samples from its own local data. In our case, the volume is based on the amount of data that is locally stored and the length of time that the smart meter has been providing data. The dataset used consists of sliding windows with a predetermined number of look-back steps. The average gradient $g_s$ is then calculated by each client $s$ using SGD, with a learning rate $\eta$. The server receives the updated models $\omega_s$ and aggregates them.

\begin{algorithm}
\caption{Federated Averaging Algorithm. $t_{max}$ is the maximum number of rounds and $M = \sum_s m_s$.\label{alg:two} }

Initialize the model in training round $t =0$\;
\While{$t \leq t_{max}$}{
Select subset $S$ of clients\;
\For{client $s$ in $S$}{
\If{$\Delta (monthlyload) > threshold$}{
$s$ obtains model $\omega_t$\;
$s$ calculates average gradient $g_s$ with SGD\;
$s$ updates local model $\omega_{t+1}^{s} \gets \omega_{t}^{s} - \eta g_s$\;
$s$ communicates updated model to server\;
}
}
server updates global model using the formula in eq (\ref{eq:update});
begin the following round $t \gets t+1$.
}
\end{algorithm}

\begin{equation}
    \label{eq:update}
    \omega_{t+1} = \sum_{s=0}^{S} \frac{m_s}{M} \omega_{t+1}^{s}
\end{equation}

\begin{figure*}[ht!]
    \centering
    \includegraphics[width=1\linewidth]{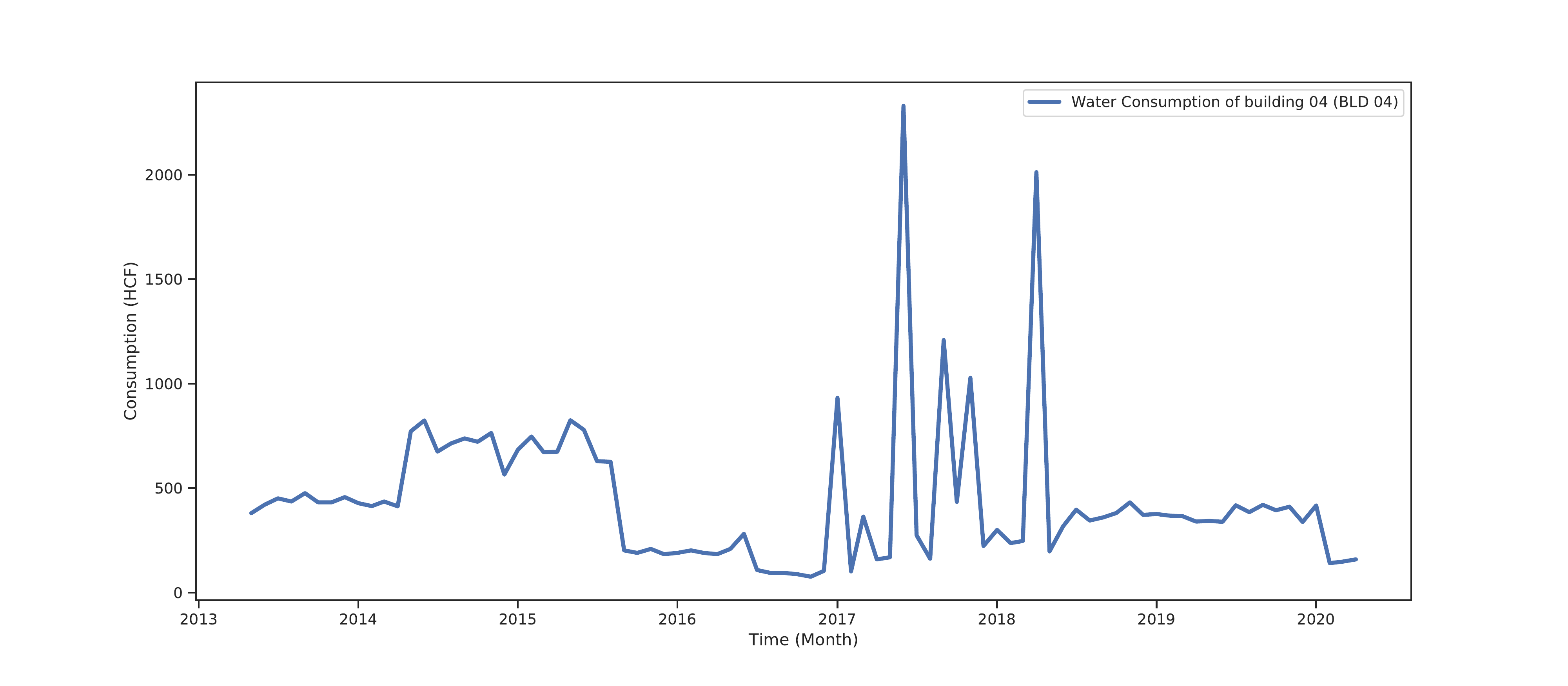}
    \caption{The distribution of water consumption data in building 04 (BLD 04).\\ \textit{ (HCF; Hundred-Cubic Foot)}}
    \label{fig:sm}
\end{figure*}

The centralized strategy might not, however, account for every user's water consumption. Personalization is one suggestion for resolving this issue. Many applications, which call for analyzing and responding to user behavior, are focused on personalization. To create a customized model for each user, the centralized model is retrained using user-specific data. This can be accomplished by retraining the model locally over a limited number of epochs using only the customer's data.\\
Since user identities are still at risk and can be ascertained through reverse engineering, federated learning poses less of a privacy threat than centralized server storage. Weight changes are processed in memory and are deleted after aggregation, whereas model updates received by each client are transient and never retained on the server. Individual weight uploads must not be reviewed or assessed according to the federated learning approach. Because the network and server cannot be trusted with precise user data, this is nevertheless more secure than server training. For billing purposes, some data must still be supplied in an aggregated form, but these data are not particularly revealing.

\subsection{Reward in the network load}

First, we establish the network load $Q_C$ for a server in centralized training in Eq. 1 and the network load in FL $Q_F$ in Eq. 2, in order to compare the reward in network load in FL to centralized training. \\
$\ell_G$ is the length of the global model, and $\ell_s$ is the length of the data that the client $s$ supplied. The number of hops between client $s$ and the server in the centralized training is denoted by $h_s$. 
\begin{equation}
  Q_C = \sum_{s=1}^M \ell_s \times h_s 
\end{equation}
and 
\begin{equation}
 Q_F = \ell_G \times \sum_{t=1}^{t_{max}} \sum_{s=1}^{S} h_{s, t}
\end{equation}
where $S$ is the total number of users in each batch, and $h_{s, t}$ is the number of hops between the client $s$ chosen in round $t$ and the server. \\
We define the reward in networking load using Eq.1 and Eq.2, as follows: 
\begin{equation}
    R = 1 - \frac{Q_F}{Q_C}
\end{equation}
\\

\section{Simulation and Results}

\subsection{Dataset Pre-Processing}

This study made use of data from the Kaggle website. Data about New York's water consumption and costs include Consumption and expense data broken down by borough and development monthly. Utility vendor and meter information is included in the data set. For over 300 buildings in the New York city, New York Water data provides water consumption and cost statistics at "monthly" intervals between 2013 and 2020.
\subsection{Numerical Results}

For deep learning models to perform the best in forecasting, hyper-parameter adjustment is crucial. However, we only evaluate the federated learning approach in this paper.\\
This section examines how personalization affects the efficiency of the models. First, we examine if localizing the model's training for the participating clients improves outcomes.\\

We chose a client (building; BLD 04) randomly from the sample of participants to show how personalization can improve predictions. The distribution of water consumption data in building 04 (BLD 04) between 2013 and 2020 is given as shown in Figure 2.\\
We have used the matching personalized models as well as the global model. Figure 3 illustrates the actual and anticipated consumption profiles between 2018 and 2020.\\
\begin{figure*}[t]
    \centering
    \includegraphics[width=1\linewidth]{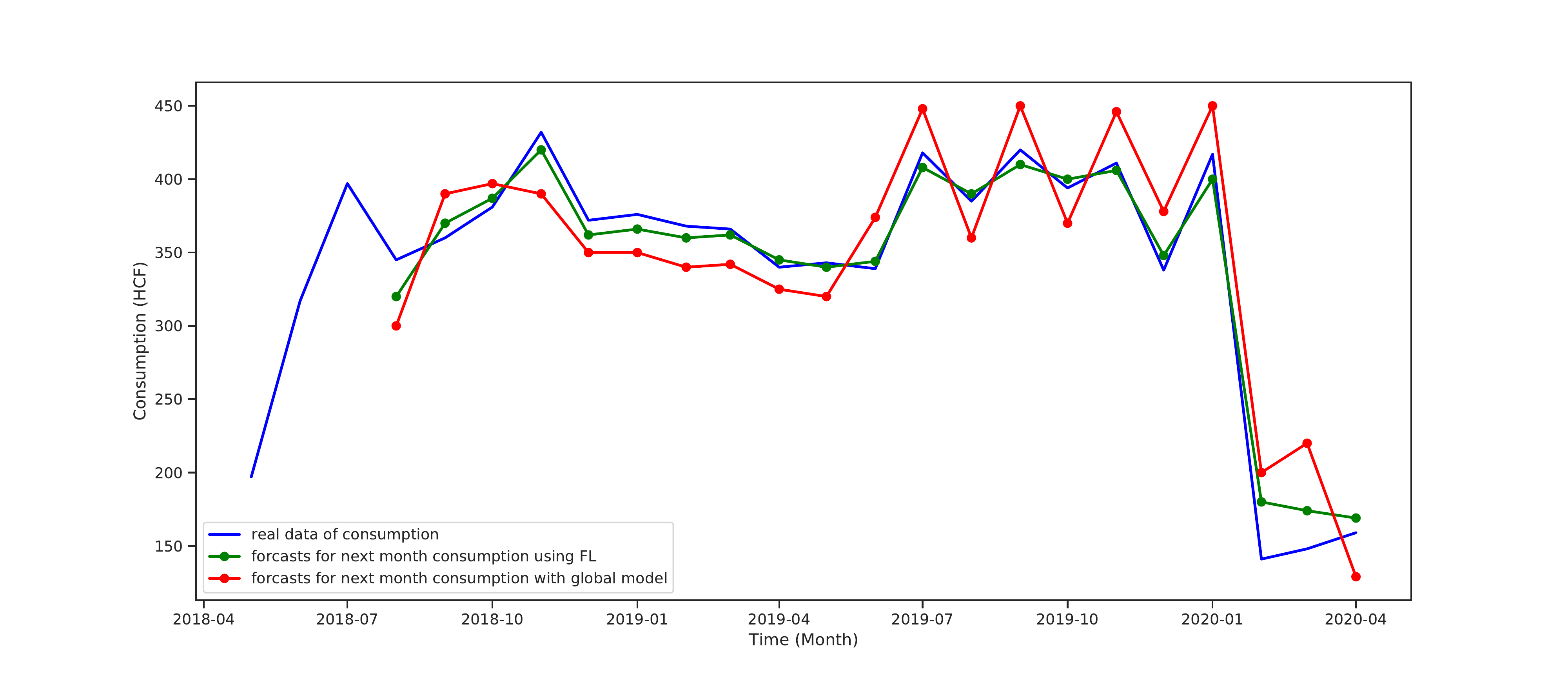}
    \caption{predictions for the client's (BLD 04) consumption for the following month for those who took part in training the global model.\\ \textit{(HCF; Hundred-Cubic Foot)}}
    \label{fig:sm}
\end{figure*}

Using just a portion of the users that make up the population, we come to the conclusion that it is possible to train robust models for the consumption profiles of that population. Retraining the model can produce a customized model that better follows the profile's characteristics and produces predictions with higher accuracy for applications with high accuracy needs. Even said, for brand-new customers lacking sufficient data for customisation, the predictions made using the global model can be a useful starting point.\\
The edge devices ability to provide local training is a necessity for the suggested approach's feasibility. The performance of new IoT devices will very certainly be compromised by training a neural network, even though they have the CPU power to execute complex machine learning models.

\section{Conclusion}
Given the stochastic nature of consumption profiles, projecting individual short-term loads is a difficult challenge. For the purpose of short-term load forecasting in smart cities, we proposed a system model in this research that uses federated learning to address issues with privacy and data diversity. Contrary to centralized techniques, federated learning in the proposed system trains models on edge devices, limiting security concerns to those specific to the device. We ran tests to assess how well centralized and customized models performed in federated contexts.
The simulation results show that it is a promising approach for developing high-performing models with a significantly reduced networking load than a centralised model, while preserving the privacy of consumption data. \\

\bibliographystyle{IEEEtran}
\bibliography{refs}

\end{document}